\documentclass[10pt,twocolumn,letterpaper]{article}

\usepackage[pagenumbers]{wacv} 

\usepackage{colortbl}
\usepackage{enumitem}
\usepackage{amsmath}
\usepackage{amssymb}
\usepackage{booktabs}
\usepackage{nicematrix}

\usepackage{bbm}
\DeclareMathOperator*{\argmin}{arg\,min}
\usepackage{multirow}
\usepackage[ruled]{algorithm2e} 

\SetAlFnt{\small}
\SetAlCapFnt{\small}
\SetAlCapNameFnt{\small}
\SetAlCapHSkip{0pt}
\newcommand{\NA}{---}
\newcolumntype{?}{!{\vrule width 0.7pt}}

\usepackage[pagebackref,breaklinks,colorlinks]{hyperref}

\usepackage[capitalize]{cleveref}
\crefname{section}{Sec.}{Secs.}
\Crefname{section}{Section}{Sections}
\Crefname{table}{Table}{Tables}
\crefname{table}{Tab.}{Tabs.}

\def\wacvPaperID{1008} 
\def\confName{WACV}
\def\confYear{2025}

\begin{document}

\title{Continual Learning of Personalized Generative Face Models with Experience Replay}

\author{Annie N. Wang\\
UNC Chapel Hill\\
{\tt\small awang13@cs.unc.edu}
\and
Luchao Qi\\
UNC Chapel Hill\\
{\tt\small lqi@cs.unc.edu}
\and
Roni Sengupta\\
UNC Chapel Hill\\
{\tt\small ronisen@cs.unc.edu}
}

\maketitle

\begin{abstract}
   We introduce a novel continual learning problem: how to sequentially update the weights of a personalized 2D and 3D generative face model as new batches of photos in different appearances, styles, poses, and lighting are captured regularly. We observe that naive sequential fine-tuning of the model leads to catastrophic forgetting of past representations of the individual's face. We then demonstrate that a simple random sampling-based experience replay method is effective at mitigating catastrophic forgetting when a relatively large number of images can be stored and replayed. However, for long-term deployment of these models with relatively smaller storage, this simple random sampling-based replay technique also forgets past representations. Thus, we introduce a novel experience replay algorithm that combines random sampling with StyleGAN's latent space to represent the buffer as an optimal convex hull. We observe that our proposed convex hull-based experience replay is more effective in preventing forgetting than a random sampling baseline and the lower bound. We introduce continual learning datasets for five celebrities, along with the evaluation framework, metrics, and visualizations to examine this problem. See our \href{https://anniedde.github.io/personalizedcontinuallearning.github.io/}{project page} for more details.
\end{abstract}

\begin{figure*}
\includegraphics[width=\textwidth]{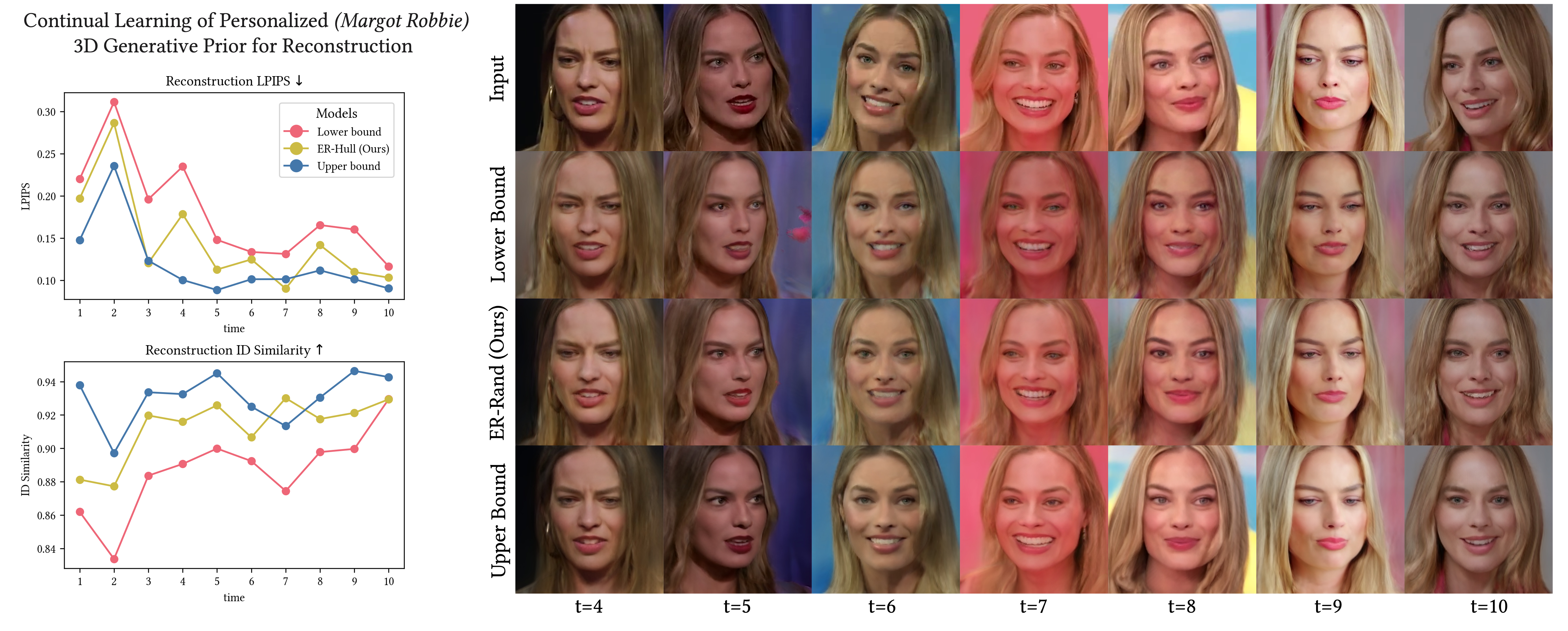}
  \caption{Open-world deployment and training of personalized generative models \cite{nitzan2022mystyle,qi2023my3dgen} are challenging as images captured at each time has limited diversity in style, appearance, and lighting conditions. Naive finetuning of the model on each timestamp  (Lower bound) leads to catastrophic forgetting, where the final model (t=10) performs poorly on test samples from previous timestamps. The Upper bound model is finetuned using all training images from all timestamps. We present an Experience Replay-based Continual Learning technique where we update a replay buffer to store the most informative images from the past as new images are captured at every timestamp. Our proposed technique, ER-Hull optimizes the buffer (size=3) as the most informative convex hull in StyleGAN's latent space.}
\label{fig:teaser}
\end{figure*}

\section{Introduction}
Generative face models \cite{karras2019style, karras2017progressive, karras2020analyzing, karras2020training, karras2021alias, niemeyer2021giraffe, schwarz2020graf, nguyen2019hologan, chan2022efficient} have proven highly effective in learning global facial priors, enabling applications such as 3D face reconstruction, novel appearance synthesis, and attribute editing. While these models can generate realistic faces, they often fail to preserve identity when reconstructing, synthesizing, or editing images of specific individuals. Recent efforts have focused on improving inversion techniques \cite{bhattarai2024triplanenet, jahanian2019steerability, shen2020interpreting} and developing personalized priors \cite{nitzan2022mystyle, qi2023my3dgen} to address this challenge. Such personalized 2D \cite{nitzan2022mystyle} and 3D \cite{qi2023my3dgen} models allow for an identity-preserving synthesis of novel appearances and semantic attribute editing, going beyond general inversion~\cite{roich2022pivotal}.

Personalization typically requires $\sim$100 images per individual \cite{nitzan2022mystyle}, encompassing diverse poses, styles, and lighting conditions to learn robust facial priors. However, accumulating such diverse photo collections can often take a significant amount of time for most users. Rather, users often capture images of themselves at the same place and time with limited stylistic variations, which also change over time due to different locations of capture or different styles of the user. Naively personalizing a generative model every time the user captures new images of themselves will lead to overfitting on a specific style and cause the model to forget previously seen facial variations resulting in poor generalization to new incoming data.


Continual learning has emerged as a promising solution to this challenge. A common technique in this field is experience replay, which involves storing the most informative past image samples in a replay buffer and integrating them with newly available image sets during model training. This method effectively reduces \textit{catastrophic forgetting} \cite{chaudhry2019tiny, rebuffi2017icarl, lopez2017gradient}, where the model loses previously learned knowledge due to overfitting on new data batches. Several experience replay algorithms \cite{chaudhry2019tiny, rebuffi2017icarl, lopez2017gradient} have been developed to select the most informative past samples. However, previous work has mainly concentrated on the task-incremental learning of conditional generative models \cite{seff2017continual, zhai2019lifelong, bartunov2018few, tao2020few}, where the generator adapts to represent different classes, or on domain-incremental learning for image classification models \cite{shin2017continual, rebuffi2017icarl, kirkpatrick2017overcoming}, a classifier learns to incrementally identify different classes. In this paper, to the best of our knowledge, we are the first to explore domain-incremental continual learning for unconditional generative models, specifically for personalized face generation.



We formulate a novel continual learning task, where at each timestamp the model is presented with a set of images of a similar style, extracted from a video recording, but the styles vary over time. We mainly explore the reply-buffer strategy for continual learning, where the model is fine-tuned at each timestamp on the newly captured image sets and the replay buffer. We propose two experience replay algorithms (`ER-Rand' and `ER-Hull') to enhance the preservation of the most informative samples in the replay buffer. We provide detailed quantitative and qualitative evaluations of personalized 2D \cite{nitzan2022mystyle} and 3D \cite{qi2023my3dgen} generative models for reconstruction and synthesis tasks, with the following observations: (1) Compared to 2D, the problem of catastrophic forgetting is more pronounced for 3D generative models. (2) For a large replay buffer size to timestamps ratio of 100\%, both random (`ER-Rand') and convex-hull optimization (`ER-Hull) based experience replay performs equally well and can match the upper bound. (3) For a small replay buffer size to timestamp ratio of 30\%, `ER-Hull' outperforms `ER-Rand' by reducing forgetting of past images by $\sim$20\% for both 2D and 3D inversion, demonstrating its effectiveness in choosing the most informative samples to reduce forgetting. 

We believe that for most practical applications a smaller buffer size compared to the number of timestamps is more storage efficient. For example, a daily update of the personalized generative model of a user for 5 years amounts to 1825 timestamps, which with 100\% buffer size to timestamp ratio will lead to $\sim$9.1GB of storage, compared to $\sim$2.7GB for 30\% buffer size to timestamp ratio (considering conservatively that each photo takes 5MB memory). However, due to limited computational resources, we limit the timestamp to $t=10$ for most of our experiments.


Our key contributions include:
\begin{enumerate}[noitemsep,topsep=0pt]
    \item A new framework for continual learning of personalized face models, enabling open-world deployment and training.
    \item Two experience replay strategies: 'ER-Rand' and 'ER-Hull', with the latter optimizing a convex hull in StyleGAN's latent space for improved performance for a smaller buffer-size to timestamp ratio.
    \item A comprehensive evaluation framework using a diverse dataset of five individuals across 10 timestamps, demonstrating the effectiveness of our approach in mitigating catastrophic forgetting.
\end{enumerate}

\section{Related Work}
\noindent \textbf{Few-Shot Personalized Generative Prior} 
Generative face models have significantly enhanced the realism of 2D facial generation \cite{karras2019style, karras2020analyzing, ho2020denoising, dhariwal2021diffusion} and facilitated the creation of 3D facial models \cite{chan2022efficient, nguyen2019hologan, niemeyer2021giraffe, or2022stylesdf, schwarz2020graf}. These advancements enable pretrained GANs to produce generalized image prior, which proves useful for tasks such as image enhancement \cite{nitzan2020face, shen2020interfacegan, patashnik2021styleclip} and semantic editing \cite{nitzan2020face, shen2020interfacegan, patashnik2021styleclip}, where images are manipulated in the GAN's latent space. However, these methods typically generate a prior across random faces, often leading to identity loss when images are projected into the latent space 

Recent works, such as MyStyle \cite{nitzan2022mystyle}, have tackled this issue by fine-tuning the generator using images of a single individual, employing an approach inspired by Pivotal Tuning \cite{roich2022pivotal} to establish a personalized prior. Similarly, for 3D generation, My3DGen \cite{qi2023my3dgen} introduces a few-shot framework that applies techniques from MyStyle \cite{nitzan2022mystyle} to learn a personalized 3D prior from EG3D \cite{chan2022efficient}. 
However, such methods operate in an offline setting, assuming that a comprehensive set of images of a person, captured across diverse poses, lighting conditions, styles, and environments, is available during training and does not require updates once training is done. This assumption may not hold in practical scenarios, where personal images typically arrive in small, consistent batches but vary over time. To address this limitation, we extend these methods to an online setting, continuously updating the model while mitigating catastrophic forgetting.

\noindent \textbf{Continual Learning and Experience Replay}
Continual learning \cite{ring1994continual} involves sequentially receiving tasks or data, acquiring new knowledge while retaining previously learned information. For a detailed overview of continual learning, readers are encouraged to consult \cite{wang2024comprehensive}. 

A major challenge in this domain is catastrophic forgetting \cite{french1999catastrophic, goodfellow2013empirical}, where a model's knowledge of earlier data deteriorates when trained on new information. Previous works \cite{chaudhry2019tiny, rebuffi2017icarl, lopez2017gradient} address forgetting through replay experience, where previously encountered training samples are stored in a replay buffer and used alongside new data during training. Various follow-up studies have proposed enhancements to experience replay through complex gradient manipulations \cite{lopez2017gradient, riemer2018learning, abrevaya2019decoupled, borsos2020coresets, Jin2020GradientbasedEO, shim2021online}. GDumb \cite{prabhu2020greedy} presents a method for greedily storing samples in memory and subsequently training a model from scratch using all available samples. This approach demonstrates superior performance compared to previously proposed algorithms \cite{lopez2017gradient, riemer2018learning, Jin2020GradientbasedEO} in their respective experimental setups. Additionally, Buzzega et al. \cite{buzzega2021rethinking} show that simple modifications to traditional rehearsal techniques can achieve performance comparable to more sophisticated methods.
These aforementioned works raise concerns about the commonly accepted assumptions, evaluation metrics, and the efficacy of various recently proposed algorithms for continual learning \cite{prabhu2020greedy}, particularly in generative models \cite{seff2017continual, zhai2019lifelong, bartunov2018few, tao2020few, sadowski2021continual}. Furthermore, there is a notable gap in the literature regarding the application of these concepts to 3D GANs and personalized 3D models, especially within the context of domain-incremental continual learning. Therefore, we propose a new problem formulation targetting this application and introduce some simple experience-replay methods to solve this new problem.

\section{Problem Setup and Methods}
\subsection{Motivation}
With the advent of generative face modeling, the problem of personalizing pretrained generative models for a particular person in a practical few-shot nature has arisen as well. MyStyle \cite{nitzan2022mystyle} and My3DGen \cite{qi2023my3dgen} tackle this problem in 2D and 3D respectively by tuning the pretrained model on a small region of the latent space and learning a personalized manifold that forms a strong personalized prior for the generator. Although their training dataset size requirements are much lower than usual generative learning methods, they do require 50-200 images captured under diverse poses, appearances, styles, and lighting conditions. These methods face the problem that real-life data does not come all at once encompassing all possible variations in style, pose, and expression. Rather images of an individual often come in batches captured over time, where each batch of images has little diversity in appearance, style, and lighting since they are captured at the same place and time. Naively fine-tuning the generative model on the newly appearing image batches leads to catastrophic forgetting of previous appearances. Also, storing all image batches over time and re-training the personalized generative model on all of them at every timestamp is highly impractical. In this paper, we examine this realistic framework of continual learning for personalizing generative models and propose two experience replay strategies to alleviate catastrophic forgetting.

\subsection{Problem Formulation}

\begin{figure}
    \centering
    \includegraphics[width=\linewidth]{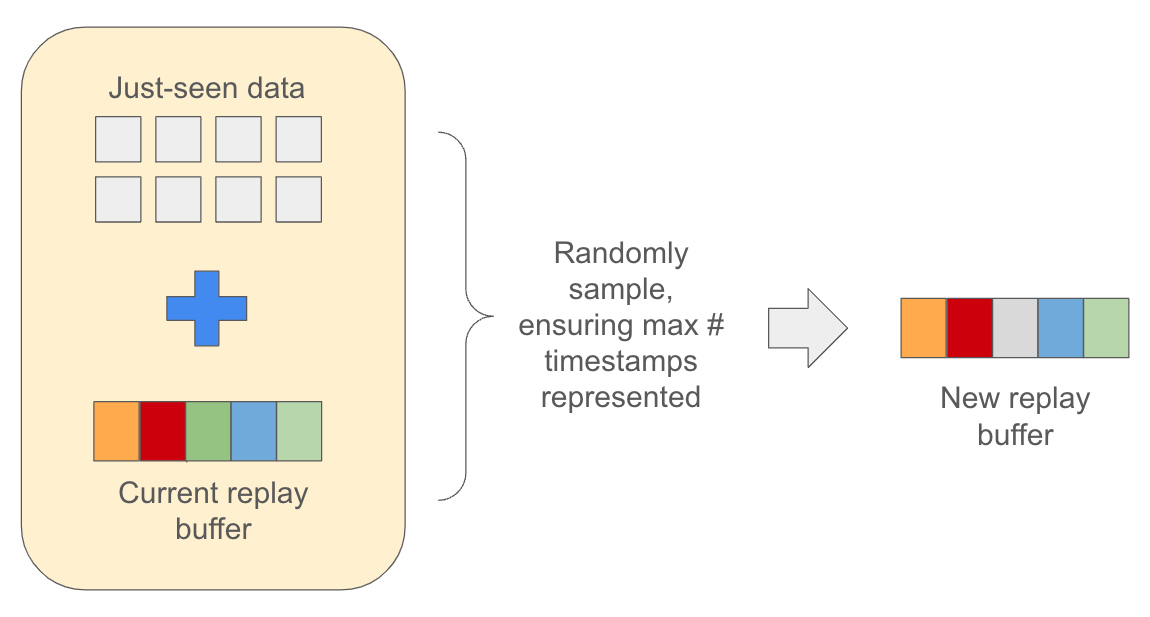}
    \caption{Diagram of the ER-Rand algorithm. We randomly sample the next replay buffer from the combination of the just-seen timestamp's data plus the replay buffer, and we only consider samples where we have the maximum possible number of timestamps represented.}
    \label{fig:ER-Rand}
\end{figure}

We consider a scenario in which we receive $T$ batches of data sequentially of a specific individual, indexed by the timestamp $t=1,2,...,T$ when it was captured. Each batch contains images of the individual captured with a similar appearance, style and lighting at same place and time, e.g. captured during a zoom call in a particular environment and day. These $T$ batches are captured at different timestamps in different environments, with varying appearance, style, and lighting across batches. Each batch is denoted as $X_t = \{x_t^1, x_t^2,...,x_t^n\}$, consisting of $n$ training images $x_t^i$ following $\mathcal{D}_t$, the data distribution for images from timestamp $t$. Our goal is to train the generative model sequentially on each data distribution such that we can best represent the combined data distribution across all timestamps.

We focus on experience replay, maintaining a small fixed-size replay buffer that stores the most informative samples from the previously seen data. Let $R_{t}$ denote the replay buffer at time $t$ that stores the $k$ most informative past images from $t=1,...,t-1$ (We define $R_1 = \emptyset$ for convenience). The goal of experience replay algorithms is to use the current batch of images $X_t$ to update the replay buffer $R_{t-1}$ to create $R_t$ and then train the generative model on $X_{t} \cup R_{t}$.


\subsection{Training Protocol}
To personalize a generator $G(\cdot;\theta)$ parameterized by $\theta$ in a continual learning fashion, we start from pretrained weights $\theta_0$ at the beginning of training. We continuously update the weights with every incoming batch of data to obtain $\theta_t$ after training on $X_t$ and the replay buffer $R_t$. For every timestamp $t$, we follow MyStyle \cite{nitzan2022mystyle} and My3DGen \cite{qi2023my3dgen} and first invert images $x_t^i \sim \mathcal{D}_t$ into the latent space of the pretrained generator $G(\cdot;\theta_0)$, obtaining latent \textit{anchors} $w_t^i \ \forall \ i=1,...,n$. Inversion is performed using a pretrained encoder \cite{richardson2021encoding}. We then optimize the network weights $\theta$, initialized at $\theta_{t-1}$ to obtain the updated weights $\theta_t$. Optimization is performed by minimizing the reconstruction objective 
\setlength{\abovedisplayskip}{3pt}
\setlength{\belowdisplayskip}{3pt}
\begin{align}
    \mathcal{L}_{rec}(G(\cdot;\theta),x_t^i, w_t^i) &= \mathcal{L}_{lpips} (G(w_t^i; \theta), x_t^i) \nonumber \\
    &+ \lambda_{L_2} ||G(w_t^i;\theta) - x_t^i||_2
\end{align}
across both the incoming batch of images $X_t$ and the replay buffer $R_t$. Formally, our loss across both sets is
\begin{equation}
    \mathcal{L}_{rec}^t = \mathbb{E}_i[ \mathcal{L}_{rec}(G, x_t^i, w_t^i) ] + \lambda_R \mathbb{E}_j[ \mathcal{L}_{rec}(G, x_{t}^{j}, w_{t}^{j}) ]
\end{equation}
where $x_{t}^{j} \in R_t \ \forall \ j = 1,...,k$ and $\lambda_R$ is a hyperparameter.

Next, we propose two basic sampling strategies to populate the replay buffer $R_t$ at every timestamp $t$.

\subsection{ER-Random and ER-Hull}
\textit{ER-Random.} Our first algorithm is a modified version of balanced reservoir sampling, shown in Fig. \ref{fig:ER-Rand}. When possible, we randomly choose a combination of currently available images from $X_t \cup R_t$ satisfying the constraint that every previously seen batch of data is represented with at least one example. However, when $t > k$, i.e. the replay buffer is not large enough to satisfy this constraint, we randomly decide whether to replace a randomly chosen example from the buffer $R_t$ with a new example from $X_t$.

This extremely simple method works well when the buffer size $k$ is large relative to the number of observed timestamps $T$, but its performance is not always optimal when the buffer size is very small compared to $T$ (for example, $T=10, k=3$). Thus, we propose a more tailored algorithm to further improve the quality of the replay buffer such that it can best capture all previously seen examples.

\begin{figure}
    \centering
    \includegraphics[width=\linewidth]{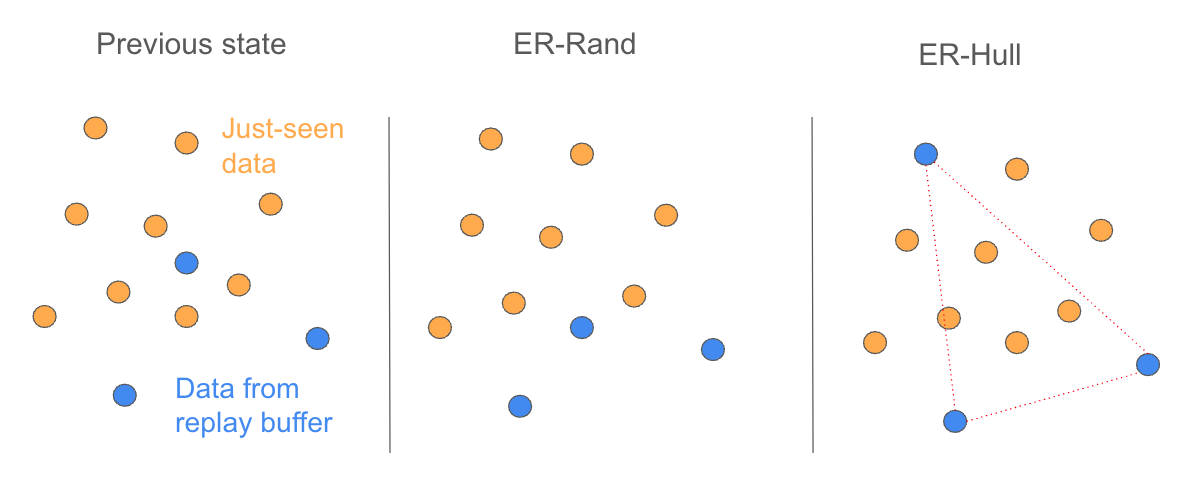}
    \caption{Illustration of ER-Hull. We perform RANSAC over many different possible replay buffers and choose the one that creates a convex hull that is "closest" to the other points, normalizing by timestamp.}
    \label{fig:ER-Hull}
\end{figure}

\textit{ER-Hull.} Our next sampling algorithm ER-Random (Fig. \ref{fig:ER-Hull}) is based on the intuition from generative face models \cite{nitzan2022mystyle, karras2020analyzing} that the latent space of StyleGAN is disentangled and the convex hull of the anchor latent codes provides a well-behaved identity-preserving prior. We hypothesize that the best image to discard from the replay buffer is the one whose latent code is closest to the convex hull of the remaining latent anchors of the replay buffer. This means the current replay buffer will be able to best preserve the information of the discarded images.




Given the data batch $X_t$ and the current replay buffer $R_{t-1}$, our goal is to choose $R_{t}$ such that the average of the distance from each currently available data batch $X^{j}_t$ to the convex hull $\textit{Hull}(R_{t})$ of the latent codes for the images in $R_{t}$ is minimized. We define the distance as:
\begin{equation}
    d(X_j, \textit{Hull}(R_{t})) = \sum_{i = 1}^n \Big(\mathbbm{1}_{[x_j^i \in X_t \cup R_{t-1}]} \cdot 
    \min_{p \in \textit{Hull}(R_{t})} |p, x_j^i|_{2} \Big),
\end{equation}
where the indicator function in the summation ensures that we only calculate the distance for samples that we have available in $X_t$ and $R_t$. Moreover, similar to ER-Random we additionally constrain the algorithm to only consider possible replay buffers that contain at least one sample from each previously seen batch of data when $k < t$ and no more than one example from each batch when $k \geq t$.

Brute-force searching over all such possible combinations of examples can be costly, so we modify the algorithm to use a RANSAC where we randomly sample at most $N$ different combinations satisfying our constraints and find the optimal replay buffer from these $N$ different options. Letting $S = \{ R_{t}^{(\ell)} \ \forall \ \ell \in 1,2,...,N\}$ denote our RANSAC-sampled set of possible buffers, we formally define our final selected replay buffer as
\begin{equation}
R_{t}^{final} = \argmin_{\ell \in \{1,2,...,N \} } \frac{\sum_{j = 1}^{t} d(X_j, \textit{Hull}(R_{t})) }{ 1 + 
\sum_{j = 1}^{t-1} \mathbbm{1}_{[X_j \cap R_{t-1} \neq \emptyset]}
}
\end{equation}
where the denominator is the number of all unique timestamps present in $X_t \cup R_{t-1}$.

\subsection{Evaluation Metrics}
We evaluate the performance of our method on both the reconstruction of test images as well as the synthesis of new images from the personalized prior. Let us define these tasks as follows:

\noindent \textbf{Reconstruction}
We maintain a held-out test set for each cluster, $X_t^{test}$, whose test examples are drawn from $\mathcal{D}_t$. We evaluate how faithful our personalized prior is through the commonly-used projection-based approach \cite{qi2023my3dgen, nitzan2022mystyle, luo2021time, menon2020pulse} of finding the best latent code in the personalized latent space that reconstructs the test image. This is done by freezing the generator and optimizing over the $W+$ latent space. We evaluate the projected images using LPIPS \cite{zhang2018unreasonable}, DISTS \cite{ding2020image}, PSNR, and ID score \cite{nitzan2022mystyle}. 

\noindent \textbf{Synthesis}
Following MyStyle \cite{nitzan2022mystyle} and My3DGen \cite{qi2023my3dgen}, synthesis is conducted for each time $t$ by sampling a latent code from the convex hull of $X_t$. For 3D synthesis, a random pose is selected as well to evaluate multiview synthesis quality. The quality and identity preservation of the synthesized image is measured through FID score \cite{heusel2017gans} and ID score \cite{nitzan2022mystyle, qi2023my3dgen}. Here, we take the maximum ID score between the synthesized image and any test image $X_t^{test}$.


\begin{table}[t]
\setlength\tabcolsep{3pt}
\centering
\caption{Continual Learning performance of personalized StyleGAN (MyStyle) in inverting an unseen test image, evaluated with Average Incremental Performance (AIP)  measured with LPIPS (lower is better) and ID similarity (higher is better) as well as Forgetting of both metrics (lower is better), scaled by $\times10$. ER-Rand and ER-Hull perform experience replay with simple random sampling and proposed convex hull optimization in StyleGAN latent space respectively.}
\label{tab:2d_reconstruction_avg}
\begin{NiceTabular}[]{cccccc}
\toprule
 \multirow{3}{*}{ \shortstack[c]{Buffer\\Size}}
 & \multirow{3}{*}{Algorithm}& \multicolumn{4}{c}{Average over 5 celebrities} \\ 
 \cmidrule(lr){3-6} 
&  & \multicolumn{2}{c}{ AIP} & \multicolumn{2}{c}{Forg. } \\ 
  \cmidrule(lr){3-4}
  \cmidrule(lr){5-6}
&\multicolumn{1}{c}{} &LPIPS &ID &LPIPS&ID\\ 

\midrule
\NA & Lower &1.17 & 8.97 & 0.64 & 1.09\\ 
 \midrule
 
\multirow{3}{*}{3}
&  KMeans-3 &1.07 & 9.25 & 0.44 & 0.68\\
&  ER-Rand&1.04 & 9.28 & 0.40 & 0.55\\
  
 & ER-Hull& \textbf{0.99} & \textbf{9.34} & \textbf{0.30} & \textbf{0.47}
  \\ \midrule
\multirow{3}{*}{5}
&  KMeans-5 & 1.00 & 9.36 & 0.29 & 0.40\\
& ER-Rand& 0.98 & 9.38 & \textbf{0.25} & 0.39\\
&  ER-Hull&  \textbf{0.98} & \textbf{9.39} & \textbf{0.25} & \textbf{0.33}\\ \midrule
\multirow{3}{*}{10}
&  KMeans-10& 0.92 & 9.44 & 0.18 & 0.26\\
&  ER-Rand& \textbf{0.91} & \textbf{9.46} & \textbf{0.15} & 0.23 \\
&  ER-Hull& \textbf{0.91} & 9.45 & 0.16 & \textbf{0.22}\\ \midrule
\NA &  Upper & 0.89 & 9.45 &\multicolumn{2}{c}{\NA}\\
\bottomrule
\end{NiceTabular}
\label{tab:2d_rec}

\end{table}

\begin{table}[h!]
\setlength\tabcolsep{3pt}
\centering
\caption{Continual Learning performance of personalized StyleGAN (MyStyle) in synthesizing novel appearance, evaluated with Average Incremental Performance (AIP) measured with FID (lower is better) and ID similarity (higher is better) as well as Forgetting of both metrics (lower is better), scaled by $\times10$. ER-Rand and ER-Hull perform experience replay with simple random sampling and proposed convex hull optimization in StyleGAN latent space respectively.}
\label{tab:2d_synth_avg}
\begin{NiceTabular}{cccccc}
\toprule
 \multirow{3}{*}{\shortstack[c]{Buffer\\Size}}
 & \multirow{3}{*}{Algorithm}& \multicolumn{4}{c}{Average over 5 celebrities} \\  
 \cmidrule(lr){3-6} 
&  & \multicolumn{2}{c}{AIP} & \multicolumn{2}{c}{Forg. } \\ 
  \cmidrule(lr){3-4}
  \cmidrule(lr){5-6}
&\multicolumn{1}{c}{} &FID &ID &FID&ID\\ 

\midrule
\NA & Lower & 124.7 & 7.48 & 96.7 & 2.64\\ 
 \midrule
 
\multirow{3}{*}{3} 
&KMeans-3 & 103.8 & 8.14 & 71.4 & 1.93\\
& ER-Rand&97.6 & 8.24 & 61.3 & 1.62\\
  
 & ER-Hull& \textbf{91.5} & \textbf{8.40} & \textbf{49.3} & \textbf{1.36}
  \\ \midrule
\multirow{3}{*}{5}
&KMeans-5 & 90.3 & 8.45 & 50.6 & 1.26\\
& ER-Rand& 87.8 & 8.52 & 46.1 & 1.12\\
&  ER-Hull&  \textbf{86.6} & \textbf{8.55} & \textbf{41.7} & \textbf{1.05} \\ \midrule
\multirow{3}{*}{10}
&KMeans-10 & 81.1 & 8.68 & 33.4 & 0.83\\
&  ER-Rand& 79.2 & 8.73 & 25.2 & 0.64\\
&  ER-Hull&\textbf{79.8} & \textbf{8.71} & \textbf{29.2} & \textbf{0.71}\\ \midrule
\NA &  Upper & 82.1 & 8.76  & \multicolumn{2}{c}{\NA}\\

\bottomrule
\end{NiceTabular}
\label{tab:2d_syn}

\end{table}%


For each task, we follow the standard practices of existing literature \cite{chaudhry2019tiny, wang2024comprehensive, lopez2017gradient} of evaluating overall performance as well as forgetting. We adopt the following measures for each aforementioned metric:

\noindent \textbf{Average Incremental Performance (AIP)} Let $a_{i,j}$ denote the performance of the model trained up until and including time $i$, evaluated for time $j$ where $j \leq i$. That is, for reconstruction we evaluate on the test set $X_j^{test}$ and for synthesis we use the convex hull $X_j$ to sample new latent codes. Then the average performance at time $t$ is $A_t = \frac{1}{t} \sum_{k =0}^t a_{t, k}$.
To further measure the overall historical performance of the model across time, we take the average incremental performance  $AIP = \frac{1}{T} \sum_{j = 1}^T A_j$.

\noindent \textbf{Forgetting} Rather than measuring overall performance, this metric measures the memory stability of generalization to previous data distributions. Let $f_j^i$ denote the forgetting on data cluster $j$ after the model is trained on data cluster $i$:
\begin{equation}
    f_j^i = \begin{cases}
        \max\limits_{l \in \{ 0, 1, ..., i-1\} } a_{l,j} - a_{i,j} \quad \text{if positive metric} \\
        \max\limits_{l \in \{ 0, 1, ..., i-1\} } a_{i,j} - a_{l,j} \quad \text{if negative metric}
    \end{cases}
\end{equation}
Then, we define the average forgetting to be $F = \frac{1}{T - 1} \sum_{j = 1}^{T-1} f_j^T$.
Intuitively, forgetting measures the memory stability of the network while the AIP measures overall performance over time.


\begin{table*}[h]
\setlength\tabcolsep{1pt}
\centering
\caption{Continual Learning performance of personalized EG3D (My3DGen) in reconstructing an unseen test image, evaluated with Average Incremental Performance (AIP)  measured with LPIPS (lower is better) and ID similarity (higher is better) as well as Forgetting of both metrics (lower is better), scaled by $\times10$. ER-Rand and ER-Hull perform experience replay with simple random sampling and proposed convex hull optimization in StyleGAN latent space respectively. Buffer size is 3.}
\label{tab:3d_reconstruction}
\centerline{
\begin{NiceTabular}{ccccccccccccccccccccc}
\toprule
 \multicolumn{1}{c}{}& \multicolumn{4}{c}{\textit{Margot}} & \multicolumn{4}{c}{\textit{Harry}} & \multicolumn{4}{c}{\textit{IU}} & \multicolumn{4}{c}{\textit{Michael}}& \multicolumn{4}{c}{\textbf{Average}}\\ 
 \cmidrule(lr){2-5} \cmidrule(lr){6-9}
 \cmidrule(lr){10-13}
 \cmidrule(lr){14-17}
 \cmidrule(lr){18-21}
  & \multicolumn{2}{c}{AIP} & \multicolumn{2}{c}{Forg. } & \multicolumn{2}{c}{AIP} & \multicolumn{2}{c}{Forg. } & \multicolumn{2}{c}{AIP} & \multicolumn{2}{c}{Forg. } & \multicolumn{2}{c}{AIP} & \multicolumn{2}{c}{Forg. } & \multicolumn{2}{c}{AIP} & \multicolumn{2}{c}{Forg. }\\ 
  \cmidrule(lr){2-3}
  \cmidrule(lr){4-5}
  \cmidrule(lr){6-7}
  \cmidrule(lr){8-9}
  \cmidrule(lr){10-11}
  \cmidrule(lr){12-13}
  \cmidrule(lr){14-15}
  \cmidrule(lr){16-17}
  \cmidrule(lr){18-19}
  \cmidrule(lr){20-21}
\multicolumn{1}{c}{} &LPIPS &ID &LPIPS&ID&LPIPS &ID &LPIPS&ID&LPIPS &ID &LPIPS&ID&LPIPS &ID &LPIPS&ID&LPIPS &ID &LPIPS&ID\\ 

\midrule
 Lower & 1.67 & 8.96 & 0.89 & 0.56 & 1.86 & 8.90 & 0.99 & 0.87 & 2.00 & 8.50 & 1.03 & 0.88 & 1.82 & 8.93 & 0.88 & 0.46 & 1.84 & 8.82 & 0.95 & 0.69\\ \midrule
  ER-Rand&\textbf{1.39} & 9.23 & 0.62 & 0.39 & 1.54 & 9.14 & 1.43 & 0.94 & 1.78 & 8.76 & 0.67 & 0.62 & 1.64 & 9.17 & 1.22 & \textbf{0.39} & 1.59 & 9.08 & 0.98 & 0.59 \\
  ER-Hull& \textbf{1.39} & \textbf{9.25} & \textbf{0.54} & \textbf{0.32} & \textbf{1.47} & \textbf{9.18} & \textbf{1.19} & \textbf{0.73} & \textbf{1.73} & \textbf{8.8} & \textbf{0.65} & \textbf{0.59} & \textbf{1.55} & \textbf{9.22} & \textbf{0.73} & 0.43 & \textbf{1.54} & \textbf{9.11} & \textbf{0.78} & \textbf{0.52}\\ \midrule
  Upper & 1.20&9.30& \multicolumn{2}{c}{\NA} &1.28&9.37& \multicolumn{2}{c}{\NA} &1.35 & 9.09 &\multicolumn{2}{c}{\NA} & 0.89 & 9.56&\multicolumn{2}{c}{\NA} &1.18 & 9.33 &\multicolumn{2}{c}{\NA}\\

\bottomrule
\end{NiceTabular}}
\label{tab:3d_rec}

\end{table*}%

\begin{table*}[h]
\setlength\tabcolsep{2pt}
\centering
\caption{Continual Learning performance of personalized EG3D (My3DGen) in synthesizing novel appearance, evaluated with Average Incremental Performance (AIP) measured with FID (lower is better) and ID similarity (higher is better) as well as Forgetting of both metrics (lower is better), scaled by $\times10$. ER-Rand and ER-Hull perform experience replay with simple random sampling and proposed convex hull optimization in StyleGAN latent space respectively. Buffer size is 3.}
\label{tab:3d_synthesis}
\centerline{
\begin{NiceTabular}{ccccccccccccccccccccc}
\toprule
 \multicolumn{1}{c}{}& \multicolumn{4}{c}{\textit{Margot}} & \multicolumn{4}{c}{\textit{Harry}} & \multicolumn{4}{c}{\textit{IU}} & \multicolumn{4}{c}{\textit{Michael}}& \multicolumn{4}{c}{\textbf{Average}}\\ 
 \cmidrule(lr){2-5} \cmidrule(lr){6-9}
 \cmidrule(lr){10-13}
 \cmidrule(lr){14-17}
 \cmidrule(lr){18-21}
  & \multicolumn{2}{c}{AIP} & \multicolumn{2}{c}{Forg. } & \multicolumn{2}{c}{AIP} & \multicolumn{2}{c}{Forg. } & \multicolumn{2}{c}{AIP} & \multicolumn{2}{c}{Forg. } & \multicolumn{2}{c}{AIP} & \multicolumn{2}{c}{Forg. } & \multicolumn{2}{c}{AIP} & \multicolumn{2}{c}{Forg. }\\ 
  \cmidrule(lr){2-3}
  \cmidrule(lr){4-5}
  \cmidrule(lr){6-7}
  \cmidrule(lr){8-9}
  \cmidrule(lr){10-11}
  \cmidrule(lr){12-13}
  \cmidrule(lr){14-15}
  \cmidrule(lr){16-17}
  \cmidrule(lr){18-19}
  \cmidrule(lr){20-21}
\multicolumn{1}{c}{} &FID &ID &FID&ID&FID &ID &FID&ID&FID &ID &FID&ID&FID &ID &FID&ID&FID &ID &FID&ID\\ 

\midrule
 Lower & 122.1 & 5.40 & 38.0 & 0.67 & 217.0 & 4.42 & 108.3 & 0.97 & 154.1 & 4.89 & 138.2 & 1.62 & 193.2 & 4.98 & 90.1 & 1.07 & 171.6 & 4.92 & 93.6 & 1.08\\ \midrule
 
  ER-Rand& 107.6 & \textbf{5.59} & \textbf{30.5} & 0.56 & 192.2 & 4.88 & 99.1 & 0.73 & 106.6 & 5.45 & 63.0 & \textbf{0.78} & 161.2 & \textbf{5.48} & 59.3 & 0.65 & 141.9 & 5.35 & 63.0 & 0.68
  \\
  
  ER-Hull& \textbf{106.2} & 5.55 & 31.2 & \textbf{0.52} & \textbf{183.1} & \textbf{4.99} & \textbf{84.6} & \textbf{0.60} & \textbf{97.6} & \textbf{5.47} & \textbf{50.6} & 0.91 & \textbf{153.6} & \textbf{5.48} & \textbf{53.8} & \textbf{0.60} & \textbf{135.1} & \textbf{5.37} & \textbf{55.0} & \textbf{0.66}
  \\ \midrule
  
  Upper & 68.6& 5.83& \multicolumn{2}{c}{\NA} &  124.7& 5.42 & \multicolumn{2}{c}{\NA} & 58.0 & 5.74 &\multicolumn{2}{c}{\NA} & 93.2 &  6.11 &\multicolumn{2}{c}{\NA} &  86.1 & 5.78 &\multicolumn{2}{c}{\NA}\\

\bottomrule
\end{NiceTabular}}
\label{tab:3d_syn}

\end{table*}%


\section{Experiments}
\subsection{Experimental Setup}

\noindent \textbf{Data}
We introduce new personal face datasets of 5 celebrities (\textit{Margot Robbie, Harry Styles, Sundar Pichai, Michael B. Jordan, IU}), each consisting of 10 batches of data (timestamps). Each batch contains 20 training images and 10 test images. Each batch contains images from video frames crawled from a single online video of the celebrity. These videos include interview videos and other short-form content uploaded online. The videos were chosen to have relatively consistent style and environment throughout, so that images from the same batch are captured with similar lighting and appearance but with variations in pose and expression, to model the real-world use case where an individual takes multiple photos of themself in the same environment. The raw frames were automatically aligned and cropped \cite{kazemi2014one} to size $512 \times 512$, and filtered to only include faces of the specified identity \cite{serengil2024lightface}. We note that due to the inherent motion in videos and the large amount of cropping, some of the images are fairly low resolution and are of lower quality than $FFHQ$ portrait images. 

\noindent \textit{Fair use disclaimer.} Our dataset is collected from YouTube videos and may contain copyrighted material. Such material is made available for research purposes only. This constitutes `fair use' under Section 107 of the Copyright Act of 1976. All rights and credit go directly to its rightful owners. No copyright infringement is intended.

\noindent \textbf{Training and Testing Details}
For the lower bound and both ER models, we train MyStyle for 1000 iterations (1 hour on 1 RTX A4500) and My3DGen for 500 interation (5 hours on 4 RTX A6000 GPUs) for each timestamp. For the upper bound, we train MyStyle for 1000 iterations and My3DGen for 2000 iterations (because it is more difficult to converge than MyStyle). For our experience replay algorithms, we set $\lambda_R = 1$ and modify the training of MyStyle and My3DGen accordingly so that 50\% of the training images at every iteration are from the replay buffer. For sampling, ER-Rand takes $\sim$ 3 seconds while ER-Hull takes 40 minutes per timestamp for $N=5000$ ransac iterations. We test replay buffer sizes of 3, 5, and 10 for 2D and buffer size 3 for 3D due to the higher compute requirement in 3D. We evaluate our results using the tasks described in section 3.6. Note that we did not use Pivotal Tuning \cite{roich2022pivotal} in our reconstruction evaluation to focus on understanding the memory and generalization power of the model without per-image tuning. We also compare our methods ER-Rand and ER-Hull to the baseline experience replay method of K-Means clustering \cite{chaudhry2019tiny} on the 2D tasks, which both have computational requirements similar to that of ER-Rand, taking $<$5 seconds to sample per identity.

\noindent \textbf{Computational complexity discussion}
We note that the training process for this experiment set up is very expensive, as for 2D, each timestamp requires 1 hour to train and 10 timestamps requires 10 hours to train. For 3D, each timestamp requires 5 hours to train and 10 timestamps requires 50 hours to train. For the actual experience replay sampling, ER-Rand takes 3 seconds for each timestamp and ER-Hull takes 40 minutes (for all buffer sizes). However, compared to the extensive training time especially for 3D experiments, the additional time required for ER-Hull is not so significant. For any given buffer size, the storage cost of both algorithms is $O(b)$ where $b$ is the buffer size.

\subsection{Evaluation on 2D Generative Models}
\label{sec:res_2d}
We first investigate the role of continual learning in open-world deployment of the personalized 2D generative model, MyStyle \cite{nitzan2022mystyle}. We mainly focus on evaluating the inversion and synthesis capability of the MyStyle model at each timestamp on heldout test images of that and all previous timestamps. For inversion, we calculate LPIPS  and ID similarity metrics. For synthesis we random sample latent codes in the convex hull defined by the buffer and the current batch of images and report FID and ID similarity metrics.

In Table \ref{tab:2d_rec} we report the Average Incremental Performance (AIP) and Forgetting for inversion tasks with LPIPS and ID similarity metric, averaged over all 5 celebrity datasets. We observe that for smaller buffer size of 3, ER-Hull improves AIP by 5\% (0.99 vs 1.04) and Forgetting by 25\% (0.3 vs 0.4) over ER-Rand in terms of LPIPS metric. For larger buffer size of 10, both ER-Rand and ER-Hull perform almost equal to the ideal Upper bound performance. 

Similarly, in Table \ref{tab:2d_syn} we report AIP and Forgetting for synthesis tasks with FID and ID similarity metric, averaged over all 5 celebrity datasets. We observe that for smaller buffer size of 3, ER-Hull improves AIP by 6.3\% (91.5 vs 97.6) and Forgetting by 19.6\% (49.3 vs 61.3) over ER-Rand in terms of FID metric. For larger buffer size of 10, ER-Hill is slightly worse than ER-Rand and only slightly worse than the ideal Upper Bound performance.

\begin{figure*}[h!]
    \centering
    \begin{subfigure}{\textwidth}
        \centering
        \includegraphics[width=0.95\textwidth]{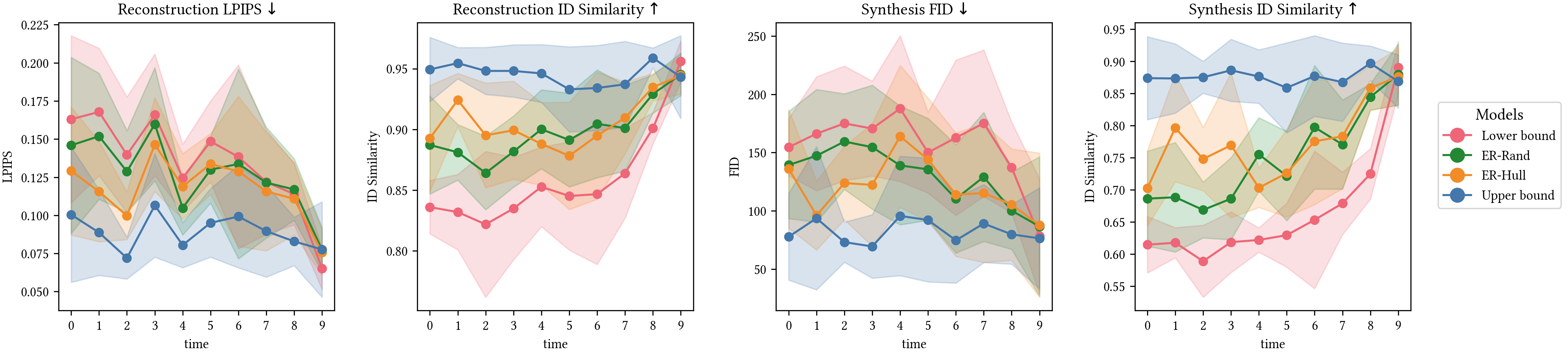}
        
    \caption{Performance of 2D generative model trained at time=10}
    \label{fig:2d_graph}
    \end{subfigure}

    \begin{subfigure}{\textwidth}
        \centering
        \includegraphics[width=0.95\textwidth]{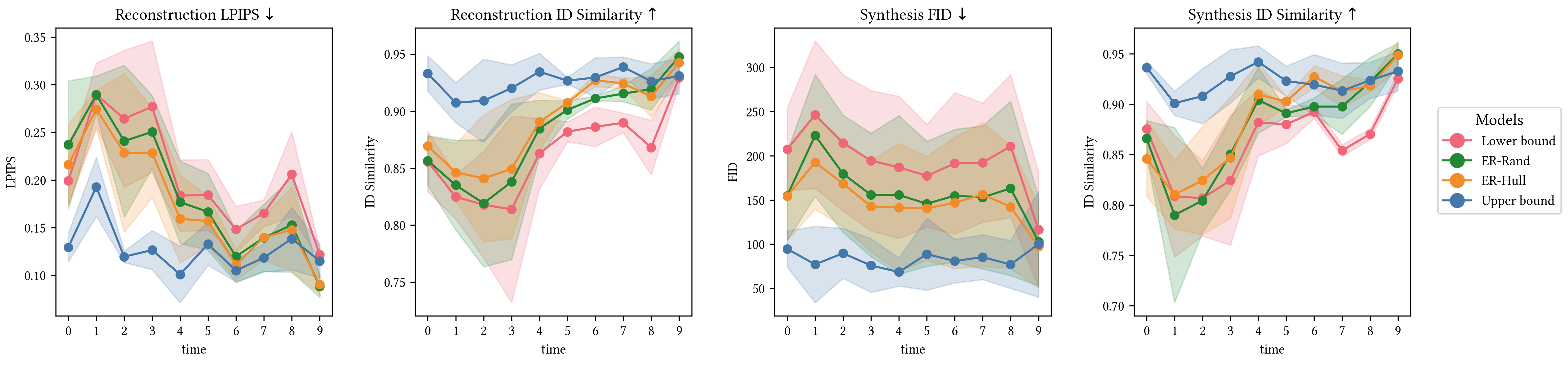}
       
    \caption{Performance of 3D generative model trained at time=10}
    \label{fig:3d_graph}
    \end{subfigure}
    
    \caption{Performance deterioration of the final personalized (a) 2D generative model and (b) 3D generative model trained at t=10 on all previous time, averaged over 5 celebrities for 2D and 4 celebrities for 3D. ER-Hull outperforms ER-Rand on earlier timestamps proving its effectiveness in reducing forgetting.}
    \label{fig:2d_3d_fig}
    
\end{figure*}




In addition, in Fig \ref{fig:2d_graph} we visualize the performance deterioration of the final model trained at timestamp 10 on all previous timestamps, averaged over all 5 celebrity datasets, for inversion with LPIPS and ID similarity and for synthesis with FID and ID similarity metrics. We only show a buffer size of 3 since that is the most challenging and ideal for long-term deployment.

\noindent \textbf{Analysis.} In summary, we observe that a larger buffer size (10) with a 1:1 ratio with the number of timestamps makes the continual learning method easier and any reasonable experience replay-based technique can perform well and closely match the ideal Upper Bound performance. In many real-life scenarios for long-term deployment, timestamps can be in $\sim$100s or $\sim$1000s, where a 1:1 ratio between buffer size and timestamp can be extremely prohibitive. For example, storing 1000 images with JPEG compression in a buffer can lead to $1000\times 5\text{ MB}=5\text{ GB}$ of memory. This is where a smaller buffer size, e.g. in a 3:10 ratio, is more practical and presents a more challenging scenario to study continual learning. With a smaller buffer size, ER-Hull can significantly reduce forgetting compared to a simple algorithm like ER-Rand.

Even though in our experiments all of the buffer sizes are trivially small due to computational limitations, the difference between a 30\% buffer size and a 50\% buffer size can be huge when the number of timesteps scales up. We performed another experiment (see Tab. \ref{tab:2d_extend_timestamps_synth}) where we extended the number of timestamps to 20 and used a buffer size of 6 (in accordance with the 30\% buffer size of ER-Rand-3 and ER-Hull-3) on a single identity (\textit{Michael B. Jordan}) due to limited resources, and found that ER-Hull still demonstrated much less forgetting than ER-Rand. This suggests that our method can be especially useful when the number of timesteps scales up significantly.

\begin{table}[h!]
\setlength\tabcolsep{3pt}
\centering
\caption{Continual Learning performance of personalized StyleGAN (MyStyle) in synthesizing novel appearance for \textit{Michael B. Jordan}, evaluated with Forgetting metrics measured with FID (lower is better) and ID similarity (higher is better, scaled by $\times10$) over 20 timestamps}
\label{tab:2d_extend_timestamps_synth}
\begin{NiceTabular}{cccc}
\toprule
 \multirow{2}{*}{\shortstack[c]{Buffer\\Size}}
 & \multirow{2}{*}{Algorithm}&\multicolumn{2}{c}{Forgetting }\\  
  \cmidrule(lr){3-4}
&\multicolumn{1}{c}{} &FID &ID \\ 

\midrule
 
\multirow{2}{*}{6} 
& ER-Rand& 50.98  & 1.5\\
  
 & ER-Hull& 29.80 & 1.0\\ 

\bottomrule
\end{NiceTabular}
\label{tab:2d_syn}

\end{table}%

\subsection{Evaluation on 3D Generative Models}
\label{sec:res_3d}

Next, we follow a similar investigation for continual learning of personalized 3D generative model (EG3D), My3DGen \cite{qi2023my3dgen}, in reconstructing an unseen test image and synthesizing novel appearance, using the same metrics as in Sec \ref{sec:res_2d}. In Table \ref{tab:3d_rec} and Table \ref{tab:2d_rec} we report AIP and Forgetting for reconstruction and synthesis tasks. We mainly focus on a buffer size of 3 since we observe that to be the most challenging scenario with large practical significance for open-world deployment for a longer time. We observe that ER-Hull is slightly better than ER-Rand for all 4 celebrities for both reconstruction and synthesis. ER-Hull improves Forgetting over ER-Rand by 20.4\% (LPIPS: 0.78 vs 0.98) for reconstruction and by 12.7\% (FID: 55 vs 63) for synthesis.

We also visualize the performance deterioration of the final model trained at timestamp 10 on all previous timestamps, averaged over all 4 celebrity datasets, for inversion with LPIPS and ID similarity and for synthesis with FID and ID similarity metrics in Fig. \ref{fig:3d_graph}. Qualitative evaluation of the final model for reconstruction and synthesis for ER-Rand and ER-Hull, compared to Lower and Upper bound performance is presented in the supplementary material.

\noindent \textbf{Analysis.} In summary, we observe that while ER-Hull is only slightly better than ER-Rand on average across all timestamps, it is significantly more effective in reducing forgetting, which is often the primary goal of a continual learning algorithm for open-world deployment. We also observe 3D tasks to be particularly more challenging than 2D tasks resulting in lower reconstruction and synthesis accuracy. This observation is supported by previous research on generative face models \cite{karras2019style,nitzan2022mystyle,chan2022efficient,or2022stylesdf,qi2023my3dgen} where 3D generation was shown to be significantly more challenging than 2D and often requires larger data with more diversity for personalization.

\section{Conclusion}

Our work is the first to tackle the problem of open-world deployment of personalized generative models. We introduce a novel problem formulation, dataset, experimental framework, evaluation metrics and visualizations to examine this problem. We introduce two experience replay-based continual learning techniques; a simple random sampling-based solution (ER-Rand) that works well for larger buffer sizes, and a more advanced one that optimizes convex hull maximization in StyleGAN latent space (ER-Hull) and works better for smaller buffer sizes. ER-Hull improves over the lower bound and closely matches the upper bound for large buffer sizes.

\noindent
\textbf{Limitations and Future Work.} While ER-Hull is better than ER-Rand for small buffer sizes, the performance with respect to the upper bound is still poor, highlighting the need for future research. Additionally, we test our model on 10 timestamps with 20 images each due to limitations in resources. This problem can be extended to significantly more timestamps ($\sim$100) to closely match real-world use cases. Lastly, the quality of the generated images dropped due to reliance on publicly available TV interviews or content videos which are often limited by resolution, motion blur, and low SNR.

\noindent \textbf{Ethical Considerations.} Access to personalized generative models trained with continual learning has the potential to reveal training images of the individual and generate unintended manipulated images. Recent research on securing deployment of ML models and detecting deep fake images will help to mitigate this risk.

{\small
\bibliographystyle{ieee_fullname}
\bibliography{egbib}
}

\end{document}